%% file: naaclhlt2019.tex
\documentclass[11pt,a4paper]{article}
\input{math_commands.tex}

\usepackage[hyperref]{naaclhlt2019}
\usepackage{times}
\usepackage{latexsym}

\usepackage{hyperref}
\usepackage{url}
\usepackage{multirow}
\usepackage[T1]{fontenc}
\usepackage{graphicx}
\usepackage{amsmath}
\usepackage{caption}
\usepackage{booktabs}
\usepackage{color}

\newcommand{\infill}[1]{\textcolor{blue}{#1}}

\aclfinalcopy 

\title{Text Infilling}

\author{
Wanrong Zhu$^{1}$,~~ Zhiting Hu$^{2,3}$,~~ Eric P. Xing$^{2,3}$\\
Peking University$^1$,~ Carnegie Mellon University$^2$,~ Petuum Inc.$^3$
}

\date{}

\begin{document}
\maketitle
\begin{abstract}
Recent years have seen remarkable progress of text generation in different contexts, such as the most common setting of generating text from scratch, and the emerging paradigm of retrieval-and-rewriting. Text infilling, which fills missing text portions of a sentence or paragraph, is also of numerous use in real life, yet is under-explored. Previous work has focused on restricted settings by either assuming single word per missing portion or limiting to single missing portion to the end of text. This paper studies the general task of text infilling, where the input text can have an arbitrary number of portions to be filled, each of which may require an arbitrary unknown number of tokens. 
We study various approaches for the task, including a self-attention model with segment-aware position encoding and bidirectional context modeling.
We create extensive supervised data by masking out text with varying strategies. Experiments show the self-attention model greatly outperforms others, creating a strong baseline for future research\footnote{Data and code are available on \url{https://github.com/VegB/Text_Infilling}}.
\end{abstract}

\section{Introduction}

Text generation spans a rich set of tasks that aim to generate natural language from input data. Popular tasks include machine translation, summarization, dialogue, and others. Previous work has made remarkable progress in text generation in various contexts.
For example, the most common setting is to generate an entire text sequence from scratch~\citep{mikolov2010recurrent,seq2seq,bahdanau2014neural}. Recent work additionally leverages retrieved reference text to help with generation~\citep{guu2017generating,weston2018retrieve}, and others~\citep{hu2017toward,shen2017style,yang2018unsupervised} generate by manipulating specific aspects of given text.

Text infilling, which fills missing text snippets of a sentence or paragraph, is also a common application in real life useful in numerous contexts, such as restoration of historical or damaged documents, contract or article writing with templates, text editing, and so forth. The counterpart application in visual domain is \emph{image inpainting} (filling missing pixels in images) which has attracted great research and industrial interest and achieved impressive results~\citep{image_inpainting,criminisi2004region,liu2018image,genrative_image_inpainting}. Text infilling, in contrast, is less explored or has been studied in simplified and more restricted settings. For example, the recent MaskGAN work~\citep{maskgan} and the sentence completion task~\citep{zweig2011microsoft} have assumed each missing portion of a sentence contains only a \emph{single} word. The assumption fails to meet the general text infilling need that each part can miss an arbitrary number of tokens and the missing word count is unknown \emph{a priori}. Other work~\citep{holtzman2018learning,fan2018hierarchical} assume the missing text are at the end of a sentence or paragraph, and continuations of the given text are generated. \citet{sun2017bidirectional} study image captioning with a single blank surrounded by known text. These studies are not directly applicable to many real scenarios where multiple portions at random positions of the text can be missing.

In this paper, we study the general task of text infilling. Consider input text where an arbitrary number of portions are missing and each portion may originally contain an arbitrary unknown number of tokens. The task aims to fill the missing portions based on the global and surrounding context, to make the text complete and meaningful. For example, given an incomplete sentence (which we call a \emph{template}) ``{\it \_\_\_\_ have a \_\_\_\_ , please .}'', the desired output could be ``{\it \underline{Can I} have a \underline{beef burger with cheddar} , please .}''. To the best of our knowledge, such general, unconstrained text infilling setting has not been studied previously. 

We make preliminary exploration of possible solutions to the task, such as the common attentional sequence-to-sequence model~\citep{bahdanau2014neural} and GAN-based approach~\citep{gan}. In particular, to better capture the global and surrounding context of the missing portions, we leverage a self-attention model~\citep{transformer} and devise a segment-aware position encoding mechanism to enable precise localization when there are multiple missing segments and varying number of missing tokens in each. 

We conduct extensive experiments in multiple concrete setups, using randomly or schematically masked text of varying number of segments and missing ratios. 
Automatic and human evaluations show the self-attention model performs reasonably well, and can serve as a strong baseline for the task in future research.

Interestingly, the concurrent work uses a similar model and training objective for text \emph{representation learning}, while focusing on text generation. It would be interesting to leverage the pre-trained model from~\citep{devlin2018bert} for the text infilling task, which we leave for future work.

\section{Related Work}
\label{related_work}
The field of text generation has undergone rapid progress in both academia and industry. This paper studies the new general setting of text infilling, which has the potential to further extend the application scope of text generation techniques in real-world tasks such as historical document restoration, article writing, text editing, etc. Deep neural networks have been widely used in many text generation tasks. Sequence-to-sequence (seq2seq)~\citep{seq2seq} with attention~\citep{bahdanau2014neural,luong2015effective} is among the most popular models. Recent efforts have also been made to apply adversarial training~\citep{gan} for text generation, among which MaskGAN~\citep{maskgan} is of particular relevance to ours. Our text infilling setting is different as it allows an arbitrary unknown number of tokens (instead of a single token) in each blank. We study a simplified GAN-based method in our setting. It would be interesting to also generalize MaskGAN and explore its performance in our task in the future. The best-performing approach in our study is based on self-attention~\citep{transformer}, resembling the Transformer encoder that encodes bi-directional context. The concurrent work of \citep{devlin2018bert} learns a text representation model with a training objective of reconstructing a randomly masked token. They also show the effectiveness of encoding bi-directional context for text modeling. Our work is independently developed, and the task of text infilling can be seen as a generalization of the random word reconstruction.

\section{Text Infilling}
\label{model}

\subsection{Problem Definition}
We consider the following problem setting: given a text template where portions of a body of text are deleted or redacted, we want to fill in the blanks
properly to produce complete, semantically coherent and meaningful text.

\begin{figure}[t]
\begin{center}
\includegraphics[width=\linewidth]{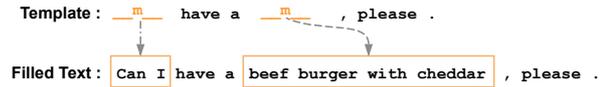}
\end{center}
\vspace{-5pt}
\caption{An example of text infilling.}
\label{fig:template}
\vspace{-5pt}
\end{figure}

Figure~\ref{fig:template} gives an example. Let \_\_$\evm$\_\_ denote a placeholder for a blank, which has masked out multiple tokens in a row. The example template has two blanks, resulting in four \emph{segments}, namely, the first blank, the snippet ``\emph{have a}'', the second blank, and the snippet ``\emph{, please .}''. An example filled text is shown in the figure.

We study the problem in a \emph{supervised} setting. That is, we assume a set of pairs including both a template and example filled text for training. Note that for each input template, the number of blanks and their positions are known, but the number of tokens to be infilled for each blank is not given. A model must decide by itself how many tokens to generate for a blank.





\subsection{Preliminary Solutions}
We explore several simple yet representative solutions that have been popularly used in other tasks, including attentional seq2seq~\citep{bahdanau2014neural}, a GAN-based model~\citep{gan}, and a method with self-attention~\citep{transformer}. All methods have similar specifications. Here we briefly describe the self-attention model adapted from~\citep{transformer}.  

We surmise a self-attention mechanism is particularly suitable for the infilling task, as it (as opposed to the sequential left-to-right attention) enables to model both the left- and right-side context of each blank, making an effective encoding the global semantics. 

The model is a simple singleton self-attention network that generates tokens in the blanks one by one. Each time when generating a token, the model (self-)attends to all other known tokens (including the tokens given in the template and the already-generated ones) and computes a distribution over the vocabulary from which the infilling token is drawn. A blank is completed when a special \texttt{<End-of-Blank>} token is generated. Then the model moves on to fill other blanks.

As the self-attention mechanism does not model position information \emph{per se}, additional positional embedding of each token is usually used~\citep{transformer}. However, in the text infilling task, as each blank can have an arbitrarily, \emph{a priori} unknown number of tokens, the conventional single-scalar position index is insufficient for uniquely localizing a token. We instead use the segment id together with the token's offset within the segment to localize each token. For example, a position index $(2, 1)$ indicates the $1$st token in the $2$nd segment, which, in the example of Figure~\ref{fig:template}, corresponds to the token ``have''. The model learns embeddings for the 2-dim position indexes.

More model details and a figure of the model architecture are presented in the appendix. The seq2seq model also generates the infilling tokens sequentially, yet conditioning on the encoded representation of the input template by the encoder. The GAN-based model adds an additional discriminator over the seq2seq to encourage global coherence. We defer more details in the appendix.

\section{Experiments}
\label{experiments}
We study the performance of the above solutions for the text infilling task. To this end, we devise diverse supervised datasets by masking out text portions with different strategies, and train the models to recover the original text.

We use LSTM RNNs for the seq2seq model, and a ConvNet as the discriminator in the GAN-based model. Same as in the self-attention model, both the seq2seq and GAN-based models also use the positional embedding as inputs. The self-attention model has 6 blocks. Please see Appendix.\ref{sec:training-details} for detailed configurations. Code are implemented with Texar~\citep{hu2018texar}, a general-purpose text generation toolkit.

\begin{table}[!t]
\resizebox{\columnwidth}{!}{%
\begin{tabular}{c  r  l  l  l  l}
\cmidrule[\heavyrulewidth]{1-6}
\#Blanks          & Metric & Template & Seq2Seq & GAN  & Self-attn \\  \cmidrule{1-6}
 \multirow{3}{*}{1} & BLEU     & 63.916   & 69.097  & 68.470  & {\bf 71.104} \\
                     & Perplexity      &    -     & 107.480 & 144.127 & {\bf 38.304} \\ 
                     & Human Eval &    -     & 1.950   & 1.775 & {\bf 2.275} \\ \cmidrule(l){2-6}

 \multirow{3}{*}{2} & BLEU      & 42.233   & 64.174  & 64.337  & {\bf 65.914} \\ 
                     & Perplexity      &      -   & 43.044  & 36.704  & {\bf 21.028} \\
                     & Human Eval &    -     &  1.838  & 1.975   & {\bf 2.188} \\ \cmidrule[\heavyrulewidth]{1-6}
\cmidrule[\heavyrulewidth]{1-6}
\#Blanks    & Metric & Template & Seq2Seq & GAN  & Self-attn \\  \cmidrule{1-6}
 \multirow{3}{*}{1} & BLEU      & 44.369   & 48.865  & 48.861  & {\bf 51.55} \\ 
                     & Perplexity      &    -     & 244.862 & 287.415 & {\bf 43.688} \\
                     & Human Eval &    -     & 1.725   &  1.863 &  {\bf 2.412} \\ \cmidrule{2-6}

 \multirow{3}{*}{2} & BLEU      & 32.498   & 42.613  & 42.535  & {\bf 44.418} \\ 
                     & Perplexity      &      -   &  99.421 & 107.558 & {\bf 32.397} \\
                     & Human Eval &    -     & 1.875   & 1.913    & {\bf 2.238} \\ \cmidrule[\heavyrulewidth]{1-6}
\end{tabular}}
\caption{Results of varying mask rates and number of blanks. The upper part of the table is the results of mask\_rate=30\%, while the lower part is the results of mask\_rate=50\%.}
\label{yelp_score_table}
\vspace{-7pt}
\end{table}

\subsection{Varying Mask Rates and \#Blanks}\label{sec:exp-varying}
We first study the impact of the mask rate (percentage of masked tokens) and the number of blanks on model performance. Intuitively, a higher mask rate and a larger number of blanks lead to a more difficult task.
%
We use a Yelp review corpus and randomly select the mask positions and lengths according to the desired mask rate and \#blanks.
The resulting dataset contains 104K/1K sentences for training/test, with a vocabulary size of 9K.

\paragraph{Quantitative and Human Evaluation} 
\label{human_eval_rules}
We use both automatic and human evaluations to compare the different models. In particular, for human evaluation, we collected generations of each of the three models on 40 randomly-selected test instances. For each test case, we randomly permutated the three generations. We then asked ten knowledgeable human annotators to rank the generations on each of the test cases. The model with a best generation received a score of 3, and the other two models received scores of 2 and 1 according to the rank, respectively.

Table~\ref{yelp_score_table} shows the results of human evaluation and automatic metrics including test-set BLEU and perplexity. As expected, with increasing mask rate and \#blanks, the model performance (BLEU and PPL) drops. 
We can see that seq2seq and GAN provide comparable performance, while the self-attention model consistently outperforms both under varying settings in terms of different metrics, showing the advantage of bi-directional global context modeling.

\paragraph{Samples} 
Table~\ref{yelp_sample_table} shows the model outputs on a test instance (See appendix for more examples). 
We can see that seq2seq and GAN fail to generate patches that fit well to the context (e.g., seq2seq: ``\emph{at appreciated by chinese food}''; and GAN: ``\emph{live right of the app}''). In contrast, the self-attention model is able to complete the template in a way that is semantically coherent and is close to the golden text.

\begin{table}[t]
\resizebox{\columnwidth}{!}{%
\begin{tabular}{@{}r  l@{}}
\cmidrule[\heavyrulewidth]{1-2}
\textbf{Template}   &  \textbf{i live \infill{\_\_m\_\_} and i was \infill{\_\_m\_\_} chinese food .}\\ \cmidrule{1-2}
Golden &  i live \infill{\underline{right down the street}} and i was \infill{\underline{craving some good}} chinese food . \\ \cmidrule{1-2}
Seq2Seq               &  i live \infill{\underline{at a ten times}} and i was \infill{\underline{at appreciated by}} chinese food . \\ \cmidrule{1-2}
GAN                &  i live \infill{\underline{right of the app}} and i was \infill{\underline{looking for}} chinese food .\\ \cmidrule{1-2}
Self-attn           &  i live \infill{\underline{in the neighborhood area}} and i was \infill{\underline{impressed with the}} chinese food . \\ \cmidrule[\heavyrulewidth]{1-2}
\end{tabular}}
\caption{Example model outputs on a Yelp test case, where the template contains two blanks and 40\% of the tokens are masked out.}
\label{yelp_sample_table}
\vspace{-7pt}
\end{table}

\subsection{Long Content Infilling}
\label{longer_content_infilling}
We next evaluate the models on their ability of infilling long content given only a few anchor words in the templates. Different from the above study of random masks, here we mask out text portions with certain strategies, mimicking different application scenarios in practice. 
%

Specifically, we created two datasets: 
(1) \textit{Grimm's Fairy Tale}~\citep{grimm_tale}
, containing 209 tales collected by the brothers Grimm. We break long sentences into shorter clauses, each of which has at least 10 but no more than 18 tokens. The resulting dataset contains 16K/3K sentences for training/test, respectively, with a vocabulary size of 7K.
For each sentence, we mask out most of the content, leaving only one noun and one verb in the template. The resulting average mask rate is 81.3\%. 
\label{grimm_dataset}
(2) NBA news adapted from~\citep{wiseman2017challenges} to simulate news sentence generation. As above, we break sentences to each have 8-16 tokens. The resulting dataset contains 21K/5K sentences for training/test, respectively, with a vocabulary size of 8K. We mask out the content and leave in each template the name of a player or a team, and the numbers (e.g., scores, \#rebounds). The resulting average mask rate is 78.1\%.

\paragraph{Quantitative and Human Evaluation}
We use the same setup as in section~\ref{sec:exp-varying} for human evaluation. With the increasing mask rate, the infilling task becomes more open-end, making BLEU score less suitable. We thus use only the test-set perplexity for automatic quantitative evaluation.
Table~\ref{lm_ppl_table} shows the results. Consistent with the above experiments, we can see the self-attention model again improves over other comparison methods on both datasets.

\begin{table}[t]
\small
\begin{center}
\begin{tabular}{@{}c  r  l  l  l@{}}
\cmidrule[\heavyrulewidth]{1-5}
Dataset & Metrics          & Seq2Seq & GAN  &  Self-attn    \\ \cmidrule{1-5}
Grimm's & Perplexity   & 10.411  &  11.784 & \textbf{9.647} \\ 
Fairy Tale & Human Eval  & 1.991   &  1.338  & \textbf{2.664} \\ \cmidrule[\heavyrulewidth]{1-5}
NBA & Perplexity   & 10.303  & 7.245  & \textbf{6.538} \\ 
Reports & Human Eval  & 1.909  & 1.818  & \textbf{2.273} \\ \cmidrule[\heavyrulewidth]{1-5}
\end{tabular}
\end{center}
\caption{Automatic and human evaluation results for long content infilling.}
\label{lm_ppl_table}
\vspace{-5pt}
\end{table}

\paragraph{Samples} 
Table~\ref{lm_sample_table} shows example outputs by the models on both datasets. We can see that in both instances, seq2seq and GAN-based model fail to generate semantically coherent and fluent patches to fill the templates. In contrast, the self-attention model tends to produce more reasonable and meaningful results (e.g., ``\emph{defeated the Philadelphia\_76ers 114-110}'' in the second instance), though there do exist unsatisfactory parts (e.g., ``\emph{the sound said}'' in the first instance).

\begin{table}[t]
\resizebox{\columnwidth}{!}{%
\begin{tabular}{@{}r  l@{}}
\cmidrule[\heavyrulewidth]{1-2}
\textbf{Template} &  \textbf{\infill{\_\_m\_\_} sound \infill{\_\_m\_\_} be \infill{\_\_m\_\_}} \\ \cmidrule{1-2}
Golden         &  \infill{\underline{if you bear it without letting a}} sound \infill{\underline{escape you , i shall}} be \infill{\underline{free}} \\ \cmidrule{1-2}
Seq2Seq           &  \infill{\underline{and}} sound \infill{\underline{the}} be \infill{\underline{and the little , and the little , and the}} \\ \cmidrule{1-2}
GAN            &  \infill{\underline{and}} sound \infill{\underline{the}} be \infill{\underline{and the , and and}} \\ \cmidrule{1-2}
 Self-attn       &  \infill{\underline{the}} sound \infill{\underline{said , i will}} be \infill{\underline{the king}} \\ \cmidrule[\heavyrulewidth]{1-2}
 
 \textbf{Template} &  \textbf{\infill{\_\_m\_\_} Toronto\_Raptors \infill{\_\_m\_\_} 114 - 110 \infill{\_\_m\_\_}} \\ \cmidrule{1-2}
Golden         &  \infill{\underline{The}} Toronto\_Raptors \infill{\underline{defeated the Detroit\_Pistons}} 114 - 110 \infill{\underline{on Sunday at ...}} \\ \cmidrule{1-2}
Seq2Seq           &  \infill{\underline{The}} Toronto\_Raptors \infill{\underline{defeated the the}} 114 - 110 \infill{\underline{on Wednesday at the Center}} \\ \cmidrule{1-2}
GAN            &  \infill{\underline{The}} Toronto\_Raptors \infill{\underline{defeated the visiting}} 114 - 110 \infill{\underline{on Friday .}} \\ \cmidrule{1-2}
 Self-attn       &  \infill{\underline{The}} Toronto\_Raptors \infill{\underline{defeated the Philadelphia\_76ers}} 114 - 110 \infill{\underline{on Friday .}} \\ \cmidrule[\heavyrulewidth]{1-2}
\end{tabular}}
\caption{Example model outputs on Grimm's Fairy Tale (upper) and NBA Reports (lower).}
\label{lm_sample_table}
\vspace{-10pt}
\end{table}

\section{Conclusion}\label{discussion}
We have studied the new task of text infilling, which aims to fill missing portions of a given sentence/paragraph. The task generalizes previous settings and permits an arbitrary number of missing portions each of which can originally have an arbitrary unknown number of tokens. We studied several models for the task, including a self-attention model with global context modeling and segment-aware position embedding. On a variety of supervised datasets, the self-attention model improved over the seq2seq and GAN-based models. Text infilling is of wide practical use in real life. We look forward to investigating more sophisticated solutions.

\clearpage
\bibliography{naaclhlt2019}
\bibliographystyle{acl_natbib}

\clearpage

\appendix

\section{More Details of Text Infilling Self-Attention Model}
\label{sec:app:model}

Here we provide detailed description of the self-attention model adapted from~\citep{transformer} for our task. Many of the specifications are the same as in~\citep{transformer}, which we include here for sake of completeness.

\subsection{Notations}
We introduce the following notations.

Let \_\_$\evm$\_\_ be a placeholder for a blank, where multiple tokens in a row are masked out. 
It is worth noticing that we use different beginning and ending token pairs to suggest the difference between the generation of a hole infilling to that of the whole sentence. Let <$\evb$$\evo$$\evb$> and <$\eve$$\evo$$\evb$> be the beginning token and ending token of each blank, while <$\evb$$\evo$$\evs$> and <$\eve$$\evo$$\evs$> mark the first and last token for the whole sentence.


For the input sequence $\vx$ = ($\vs_1$, $\vs_2$, ..., $\vs_n$), $\vs_i$ refers to the $i_{th}$ input segment.
Let $\evx_{(i, j)}$ denote the $j_{th}$ token in the $i_{th}$ input segment $\vs_i$,  $\vs_i$ can be represent as ($\evx_{(i, 1)}$, $\evx_{(i, 2)}$, ..., $\evx_{(i, o_i)}$).
The input sequence may also be given as $\vx$ = ($\evx_{(1, 1)}$, $\evx_{(1, 2)}$, ..., $\evx_{(1, o_1)}$, $\evx_{(2, 0)}$, $\evx_{(2, 1)}$, ..., $\evx_{(2, o_2)}$, ..., $\evx_{(n, 1)}$, $\evx_{(n, 2)}$, ..., $\evx_{(n, o_n)}$).

Let $\vx_{template_i}$ denote the template sequence that is attended to fill in the blank whose $seg\_id$ is $\evi$.
We use $\vs'_i$ to refer to the filled-in segment for the blank with \textit{seg\_id = i} while $\evx'_{(i, j)}$ denotes a token in it.
Finally, let $\sM$ be the set that contains all the blanks' \textit{seg\_id}.

\subsection{Approach}

\begin{figure*}[t]
\begin{center}
\includegraphics[width=0.9\linewidth]{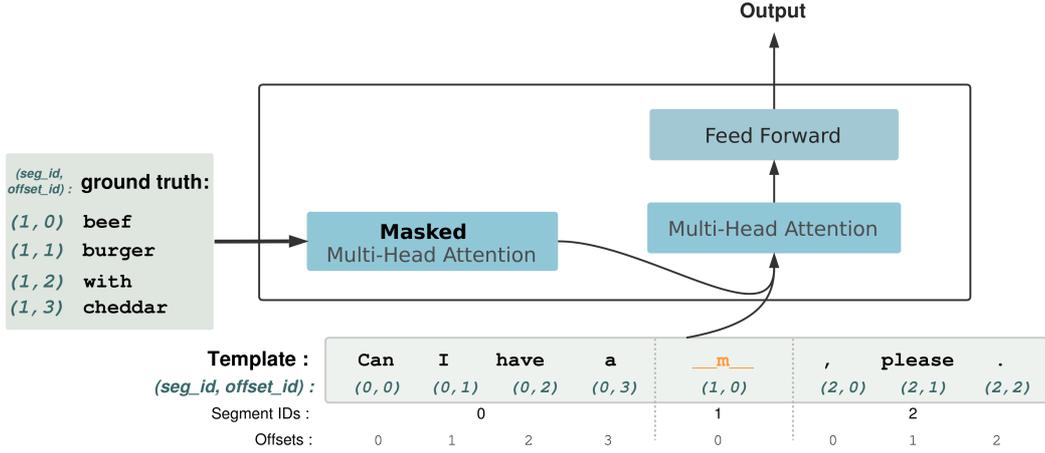}
\end{center}
\vspace{-10pt}
\caption{The overall structure of Self-attn. This figure depicts the training process. The decoder will attend to the template at each position, conditioning on the template together with what has been filled in the template. During inference, the input will not go through the masked multi-head attention layer.}
\label{fig:model}
\end{figure*}

Figure~\ref{fig:model} depicts the overall architecture of our model. 
The basis for our model is a multi-head self-attention token decoder, which fits the task of infilling as it is able to condition on information from both the past and the future. Our implementation replicates~\citep{transformer}. 

\subsubsection{Template}




\paragraph{Update Template} 
After filling in each blank, we update the template by replacing the specific placeholder \_\_$\evm$\_\_ into corresponding segment. 

Suppose segment $\evi$ and segment $\evj$ in $\vx$ $(\evi < \evj)$ are masked out in the template. 
Thus, the initial template $\vx_{template} = ( \vs_1, ...,  \vs_{i-1}, \_\_ \evm\_\_,  \vs_{i+1}, ...,  \vs_{j-1}, \_\_ \evm\_\_,  \vs_{j+1}, ...,  \vs_n)$. During training, after generating the $\evi_{th}$ segment, the ground truth $\vs_{i}$ will be filled back into the template, and template will be updated into $
 \vx_{template_{j}}
= ( \vs_1, ...,  \vs_{i-1},  \vs_{i},  \vs_{i+1}, ...,  \vs_{j-1}, \_\_ \evm\_\_,  \vs_{j+1}, ...,  \vs_n).
$
During testing, the inference segment $\vs'_{i}$ will be filled back into the template, and the new template will be $
 \vx_{template_{j}}
= ( \vs_1, ...,  \vs_{i-1},  \vs'_{i},  \vs_{i+1}, ...,  \vs_{j-1}, \_\_ \evm\_\_,  \vs_{j+1}, ...,  \vs_n)$.
The decoder will attend to the updated template $\vx_{template_{j}}$ when filling in next blank, whose \textit{seg\_id} is $\evj$.

\subsubsection{Position Encoding}
\label{positional_embedding}

Since the Self-attn architecture based solely on attention mechanism and thus contains no recurrence or convolution, we need to inject additional information about the relative or absolute position of the tokens in the sequence.

As can be seen in Figure \ref{fig:model}, the location of each token in the template can be uniquely determined by its segment number \textit{seg\_id} and the offset in that segment, which we denote as \textit{offset\_id}.
As in original Transformer~\citep{transformer}, we use sine and cosine functions of different frequencies as positional embedding:
\begin{equation*}
\small
\begin{split}
&PE_{(pos, 2i)} = sin(pos / 10000^{2i / d_{model}})\\
&PE_{(pos, 2i+1)} = cos(pos / 10000^{2i / d_{model}}),
\end{split}
\end{equation*}
where \textit{i} is the dimension and $pos = seg\_id * base + offset\_id$ is the unique position index for each token given by $(seg\_id, offset\_id)$ and a self-defined integer $base$.

The positional embeddings have the same dimension \textit{$d_{model}$} as the word embeddings, ensuring that the two can be summed. The sum of the positional embeddings and the word embeddings for the input token sequence will be used as input for the Transformer.

\subsubsection{Applications of Attention}
\label{application_of_attention}

As proposed by~\cite{transformer}, an attention function maps a query and a set of key-value pairs to an output, where the query, keys, values, and output are all vectors. The input consists of queries and keys of dimension $\evd_k$, and values of dimension $\evd_v$. We pack a set of queries, keys and values into matrix $\mQ$, $\mK$ and $\mV$ representatively to compute the attention function simultaneously. The attention function is given by:
\begin{center}
\small
$Attention(\mQ, \mK, \mV) = softmax(\frac{ \mQ \mK^T}{\sqrt{ \evd_k}})$
\end{center}
\normalsize

Multi-head attention mechanism projects queries, keys and value to different representation subspaces and calculates corresponding attention. The attention function outputs are concatenated and projected again before giving the final output. Multi-head attention allows the model to attend to multiple features at different positions.

In this work, the multi-head attention is used in the following two ways: 
(1) The decoder contains self-attention layers where the keys, values and queries come from the output of the previous layer in the decoder. This allows the decoder to attend to all previous positions and make use of local information during infilling.
(2) In "template-decoder attention" layers, the queries come from the previous decoder layer, and the template embeddings are used as memory keys and values. This makes sure the decoder can attend to all positions in the template and capture global semantic information while filling each blank.

\subsubsection{Training}

\paragraph{Objective} 
In the infilling process, the decoder will fill in the blanks one by one. For the infilling of each segment, the decoder fills in the missing token auto-regressively, conditioning on the template together with what has been filled in the template. To fill the blank with \textit{seg\_id} = $\evi$, the objective is to minimized the following cross-entropy loss:
\begin{multline*}
\small
\mathcal{L}_i ( \evx'_{(i, 0)},  \evx'_{(i, 1)}, ...,  \evx'_{(i, o_i)} |  \vx_{template_i}) \\
= - \log \prod_{j=0}^{o_i}  P(\evx'_{(i, j)} | \evx'_{(i, 0)}, ..., \evx'_{(i, j-1)}, \vx_{template_i}) \\
i \in \sM. 
\end{multline*}
The loss $\mathcal{L}$ for each infilling sentence is the sum of the cross-entropy loss for each infilling blank: 
\begin{center}
\small
$\mathcal{L} = \sum  \mathcal{L}_i, i \in \sM.$
\end{center}
\normalsize

\paragraph{Optimizing} 
We use Adam optimizer~\citep{kingma2014adam} with $\beta_1$ = 0.9, $\beta_2$ = 0.997 and $\epsilon$ = $10^{-9}$. We follow the setting in~\citep{transformer} and linearly increase the learning\_rate for the first \textit{warmup\_steps} training steps, then decrease the learning\_rate  proportionally to the inverse square root of the step number. 
We set \textit{const} = 0.3 and \textit{warmup\_step} = 10000.
\begin{center}
\small
$ learning\_rate = const * \frac{1}{\sqrt{d_{model}}} * min(\frac{step\_num}{(\sqrt{warmup\_step})^3}, \frac{1}{\sqrt{step\_num}}) $
\end{center}

\section{Training Details}\label{sec:training-details}

\subsection{Model Parameters}
\paragraph{Seq2Seq model}
The sum of template's word embedding and its positional embedding is given to the encoder. We start a loop and fill in one blank at a time. During training, the ground truth of the blank is provided to the decoder for teacher forcing. During inference, however, we only feed the special token of <$\evb$$\evo$$\evb$>(begin-of-blank) to the decoder. 

We update the template after filling in a blank and use the new template to assist the infilling of next blank.

\begin{itemize}
  \item word\_embedding\_size = 400
  \item Encoder: UnidirectionalRNNEncoder
    \begin{itemize}
      \item cell\_type = LSTM
      \item num\_units = 1600
      \item dropout\_rate = 10\%
      \item layer\_num = 1
    \end{itemize}
  \item Decoder: BasicPositionalRNNDecoder
    \begin{itemize}
      \item cell\_type = LSTM
      \item num\_units = 1600
      \item dropout\_rate = 10\%
      \item layer\_num = 1
    \end{itemize}
\end{itemize}

\paragraph{GAN-based model}
The generator is the same with Seq2Seq model.
The discriminator is trained to tell apart from the generated infilling and the ground truth for each blank along with the training of the generator. The classification result on the generated infilling is treated as the reward and is used to update the generator.

\begin{itemize}
  \item word\_embedding\_size = 400
  \item Generator: The same with Seq2Seq model
  \item Discriminator: Conv1DClassifier
    \begin{itemize}
      \item kernel\_size = [3, 4, 5]
      \item filters = 128
      \item dropout\_rate = 50\%
      \item num\_dense\_layers = 0
    \end{itemize}
\end{itemize}

\paragraph{Self-attn model}
The template is given to the Transformer Decoder as reference for future infilling. During training, the ground truth of the blank is provided to the decoder for teacher forcing. During inference, however, we only feed the special token of <$\evb$$\evo$$\evb$>(begin-of-blank) to the decoder.

\begin{itemize}
    \item word\_embedding\_size = 400
    \item Decoder: TemplateTransformerDecoder
    \begin{itemize}
      \item embedding\_dropout\_rate = 10\%
      \item attention\_dropout\_rate = 10\%
      \item residual\_dropout\_rate = 10\%
      \item position\_embedder: sinusoids embedding
      \item num\_blocks = 6
      \item num\_attention\_head = 8
    \end{itemize}
\end{itemize}

\subsection{Training Process}
\subsubsection{Training Parameters}
\begin{itemize}
    \item batch\_size = 200
    \item training\_epoch = 150
\end{itemize}

\section{Other Experiments}
\subsection{Varying Mask Rates and Segments}

In this section, we display the quantitative and human evaluations results when removing 30\%, 40\% and 50\% of the tokens in the template. With the same mask rate, we test the generation process with templates containing one or two blanks. 

Results are listed in table~\ref{app_yelp_score_table}.

\begin{table*}[h!]
\small
\center
\begin{tabular}{ r  r  r  l  l  l  l}
\cmidrule[\heavyrulewidth]{1-7}
         Mask rate     & \#Blanks        & Metric & Template & Seq2Seq & GAN  & Self-attn \\ \cmidrule{1-7}
\multirow{6}{*}{30\%}  &  \multirow{3}{*}{1} & BLEU Score     & 63.916   & 69.097  & 68.470  & {\bf 71.104} \\
                       &                      & Perplexity      &    -     & 107.480 & 144.127 & {\bf 38.304} \\ 
                       &                      & Human Eval &    -     & 1.950   & 1.775 & {\bf 2.275} \\ \cmidrule(l){2-7}

                       &  \multirow{3}{*}{2} & BLEU Score      & 42.233   & 64.174  & 64.337  & {\bf 65.914} \\ 
                       &                      & Perplexity      &      -   & 43.044  & 36.704  & {\bf 21.028} \\
                       &                      & Human Eval &    -     &  1.838  & 1.975   & {\bf 2.188} \\ \cmidrule[\heavyrulewidth]{1-7}
\multirow{6}{*}{40\%}  &  \multirow{3}{*}{1} & BLEU Score      & 56.838   & 61.309  & 61.778  & {\bf 63.543} \\ 
                       &                      & Perplexity      &    -     & 202.714 & 230.569 & {\bf 44.864} \\
                       &                      & Human Eval &    -     & {\bf 2.075 }  &  1.865  &  2.055 \\ \cmidrule(l){2-7}

                       &  \multirow{3}{*}{2} & BLEU Score      & 38.279   & 55.460 & 55.326   & {\bf 59.192} \\ 
                       &                      & Perplexity      &      -   & 59.877 &  70.195  & {\bf 25.914} \\
                       &                      & Human Eval &    -     &  2.005  &  1.900  &  {\bf 2.045} \\ \cmidrule[\heavyrulewidth]{1-7}
\multirow{6}{*}{50\%}  &  \multirow{3}{*}{1} & BLEU Score      & 44.369   & 48.865  & 48.861  & {\bf 51.55} \\ 
                       &                      & Perplexity      &    -     & 244.862 & 287.415 & {\bf 43.688} \\
                       &                      & Human Eval &    -     & 1.725   &  1.863 &  {\bf 2.412} \\ \cmidrule{2-7}

                       &  \multirow{3}{*}{2} & BLEU Score      & 32.498   & 42.613  & 42.535  & {\bf 44.418} \\ 
                       &                      & Perplexity      &      -   &  99.421 & 107.558 & {\bf 32.397} \\
                       &                      & Human Eval &    -     & 1.875   & 1.913    & {\bf 2.238} \\ \cmidrule[\heavyrulewidth]{1-7}
\end{tabular}
\caption{Quantitative and human evaluations for different mask rates and number of segments.}
\label{app_yelp_score_table}
\end{table*}

\subsection{Longer Content Infilling}

In this section, we display more examples for  infilling tasks on longer contents.

Firstly, we conduct experiments on Grimm Tales, revealing a noun and a verb as anchoring words in the template while masking out the rest. Table \ref{app_grimm_lm_sample_table} provides two examples for comparison.

\begin{table*}[h!]
\small
\center
\begin{tabular}{ r  l}
\cmidrule[\heavyrulewidth]{1-2}
\textbf{Template} &  \textbf{\infill{\_\_m\_\_} sound \infill{\_\_m\_\_} be \infill{\_\_m\_\_}} \\ \cmidrule{1-2}
Ground Truth         &  \infill{\underline{if you bear it without letting a}} sound \infill{\underline{escape you , i shall}} be \infill{\underline{free}} \\ \cmidrule{1-2}
Seq2Seq           &  \infill{\underline{and}} sound \infill{\underline{the}} be \infill{\underline{and the little , and the little , and the}} \\ \cmidrule{1-2}
GAN            &  \infill{\underline{and}} sound \infill{\underline{the}} be \infill{\underline{and the , and and}} \\ \cmidrule{1-2}
 Self-attn       &  \infill{\underline{the}} sound \infill{\underline{said , i will}} be \infill{\underline{the king}} \\ \cmidrule[\heavyrulewidth]{1-2}
\textbf{Template} &  \textbf{\infill{\_\_m\_\_} laid \infill{\_\_m\_\_} water \infill{\_\_m\_\_}} \\ \cmidrule{1-2}
Ground Truth         &  \infill{\underline{and when she had finished , she}} laid \infill{\underline{it down at the}} water \infill{\underline{'s edge .}} \\ \cmidrule{1-2}
Seq2Seq           &  \infill{\underline{and}} laid \infill{\underline{the}} water \infill{\underline{, and the little , and the little , and the}} \\ \cmidrule{1-2}
GAN            &  \infill{\underline{and}} laid \infill{\underline{the}} water \infill{\underline{and the , and and the}} \\ \cmidrule{1-2}
 Self-attn       &  \infill{\underline{and}} laid \infill{\underline{the}} water \infill{\underline{in the midst of the forest}} \\ \cmidrule[\heavyrulewidth]{1-2}
\end{tabular}
\caption{Examples for language models with anchor words on Grimm Tales.}
\label{app_grimm_lm_sample_table}
\end{table*}

We also conducted experiments on NBA reports. For each template, we use the player name or team name as well as a number related phrase as anchoring words. Table \ref{app_nba_lm_sample_table} lists two examples.

\begin{table*}[h!]
\small
\center
\begin{tabular}{ r  l}
\cmidrule[\heavyrulewidth]{1-2}
\textbf{Template} &  \textbf{\infill{\_\_m\_\_} Toronto\_Raptors \infill{\_\_m\_\_} 114 - 110 \infill{\_\_m\_\_}} \\ \cmidrule{1-2}
Ground Truth         &  \infill{\underline{The}} Toronto\_Raptors \infill{\underline{defeated the Detroit\_Pistons}} 114 - 110 \infill{\underline{on Sunday at the Air Canada}} \\ \cmidrule{1-2}
Seq2Seq           &  \infill{\underline{The}} Toronto\_Raptors \infill{\underline{defeated the the}} 114 - 110 \infill{\underline{on Wednesday at the Center}} \\ \cmidrule{1-2}
GAN            &  \infill{\underline{The}} Toronto\_Raptors \infill{\underline{defeated the visiting}} 114 - 110 \infill{\underline{on Friday .}} \\ \cmidrule{1-2}
 Self-attn       &  \infill{\underline{The}} Toronto\_Raptors \infill{\underline{defeated the Philadelphia\_76ers}} 114 - 110 \infill{\underline{on Friday .}} \\ \cmidrule[\heavyrulewidth]{1-2}
\textbf{Template} &  \textbf{\infill{\_\_m\_\_} Bojan \infill{\_\_m\_\_} 30 minutes \infill{\_\_m\_\_}} \\ \cmidrule{1-2}
Ground Truth         &  Bojan \infill{\underline{Bogdonavic was not far behind , scoring 22 points in}} 30 minutes \infill{\underline{off}} \\ \cmidrule{1-2}
Seq2Seq           &  Bojan \infill{\underline{led the way with with points points}} 30 minutes \infill{\underline{, while}} \\ \cmidrule{1-2}
GAN            &   Bojan \infill{\underline{was second on the team , totaling 19 points ,}} 30 minutes \infill{\underline{,}} \\ \cmidrule{1-2}
 Self-attn       &  Bojan \infill{\underline{led the way with 20 points in}} 30 minutes \infill{\underline{in the fourth quarter}} \\ \cmidrule[\heavyrulewidth]{1-2}
\end{tabular}
\caption{Examples of the NBA reports for language models with anchor words.}
\label{app_nba_lm_sample_table}
\end{table*}

\subsection{Preposition Infilling}
\paragraph{Dataset}
In this dataset, we also train the model on Grimm dataset, sentence number and vocabulary size are the same with section \ref{grimm_dataset}. 

We mask out preposition (e.g., \emph{in}, \emph{at}, \emph{on}, etc) and article (e.g., \emph{a}, \emph{an}, \emph{the}, etc) words in the corpus. Each template contains three blanks. The average mask rate is 20.9\%. Empty masks that remove nothing will be added to the template if there are less than three segments that satisfy such masking rules.

%

\paragraph{Samples} 
Table~\ref{app_grimm_prep_sample_table} provides an example of the preposition infilling task. seq2seq and GAN are prone to make grammatical mistakes (e.g., seq2seq: ``\emph{saw at one}''; and GAN: ``\emph{the old woman went for, }''), which indicates that these two rnn-based generative models failed to grasp the rules of using prepositions. Our model learns the rules and generates prepositions that fit into the template.

\begin{table*}[h!]
\small
\center
\begin{tabular}{ r  l}
\cmidrule[\heavyrulewidth]{1-2}
\textbf{Template}              &  \textbf{\infill{\_\_m\_\_} old woman went \infill{\_\_m\_\_} , but saw \infill{\_\_m\_\_} one on the stairs} \\ \cmidrule{1-2}
Ground Truth & \infill{\underline{the}} old woman went \infill{\underline{out}} , but saw \infill{\underline{no}} one on the stairs \\ \cmidrule{1-2}
Seq2Seq               & \infill{\underline{the}} old woman went \infill{\underline{with}} , but saw \infill{\underline{at}} one on the stairs \\ \cmidrule{1-2}
GAN                & \infill{\underline{the}} old woman went \infill{\underline{for}} , but saw \infill{\underline{no}} one on the stairs\\ \cmidrule{1-2}
 Self-attn           & \infill{\underline{the}} old woman went \infill{\underline{in}} , but saw \infill{\underline{that}} one on the stairs \\ \cmidrule[\heavyrulewidth]{1-2}
\end{tabular}
\caption{An example from the Grimm Tales data where prepositions are masked out.}
\label{app_grimm_prep_sample_table}
\end{table*}

\end{document}

%% file: math_commands.tex
\usepackage{amsmath,amsfonts,bm}

\def\eqref#1{equation~\ref{#1}}

\def\1{\bm{1}}

\def\vs{{\bm{s}}}

\def\vx{{\bm{x}}}

\def\evb{{b}}

\def\evd{{d}}
\def\eve{{e}}

\def\evi{{i}}
\def\evj{{j}}

\def\evm{{m}}

\def\evo{{o}}

\def\evs{{s}}

\def\evx{{x}}

\def\mK{{\bm{K}}}

\def\mQ{{\bm{Q}}}

\def\mV{{\bm{V}}}

\DeclareMathAlphabet{\mathsfit}{\encodingdefault}{\sfdefault}{m}{sl}
\SetMathAlphabet{\mathsfit}{bold}{\encodingdefault}{\sfdefault}{bx}{n}

\def\sM{{\mathbb{M}}}

